\documentclass[letterpaper, 10 pt, conference]{ieeeconf}  

\IEEEoverridecommandlockouts                              

\usepackage{graphicx}
\usepackage{booktabs}
\usepackage{amsfonts}
\usepackage{amsmath}
\usepackage{amssymb}
\usepackage{mathtools}
\usepackage{adjustbox}
\usepackage[table]{xcolor}
\usepackage{microtype}
\usepackage{color}
\usepackage{multirow}
\usepackage{makecell}
\usepackage{algorithmic}
\usepackage{algorithm}
\usepackage{etoolbox,siunitx}
\let\labelindent\relax 
\usepackage{enumitem}
\usepackage{hyperref}

\title{\LARGE \bf
What Really Matters for Robust Multi-Sensor HD Map Construction?
}

\author{Xiaoshuai Hao$^{1}$, Yuting Zhao$^{2}$, Yuheng Ji$^{2}$, Luanyuan Dai$^{3}$, Peng Hao$^{4,*}$\thanks{*Corresponding author.}\\ Dingzhe Li$^{4}$, Shuai Cheng$^{5}$, Rong Yin$^{6}$
\thanks{$^{1}$Beijing Academy of Artificial Intelligence.} 
\thanks{$^{2}$Institute of Automation, Chinese Academy of Science.}
\thanks{$^{3}$Nanjing University of Science and Technology.}
\thanks{$^{4}$Samsung R\&D Institute China–Beijing.}
\thanks{$^{5}$China North Artificial Intelligent \& Innovation Research Institute.}
\thanks{$^{6}$Institute of Information Engineering, Chinese Academy of Sciences.}
\thanks{E-mail:xshao@baai.ac.cn; peng1.hao@samsung.com.}
}

\begin{document}

\maketitle
\thispagestyle{empty}
\pagestyle{empty}

\begin{abstract}

High-definition (HD) map construction methods are crucial for providing precise and comprehensive static environmental information, which is essential for autonomous driving systems. While Camera-LiDAR fusion techniques have shown promising results by integrating data from both modalities, existing approaches primarily focus on improving model accuracy, often neglecting the robustness of perception models—a critical aspect for real-world applications.
In this paper, we explore strategies to enhance the robustness of multi-modal fusion methods for HD map construction while maintaining high accuracy. 
We propose three key components: data augmentation, a novel multi-modal fusion module, and a modality dropout training strategy. These components are evaluated on a challenging dataset containing 13 types of multi-sensor corruption. 
Experimental results demonstrate that our proposed modules significantly enhance the robustness of baseline methods. Furthermore, our approach achieves state-of-the-art performance on the clean validation set of the NuScenes dataset.
Our findings provide valuable insights for developing more robust and reliable HD map construction models, advancing their applicability in real-world autonomous driving scenarios.
Project website: \url{https://robomap-123.github.io/}.
\end{abstract}

\section{Introduction}
\label{sec:intro}

High-definition (HD) map construction is a critical task for autonomous driving systems, providing rich semantic and geometric road information essential for localization, perception, and path planning. HD maps capture key details such as lane boundaries and road markings, which are vital for the precise operation of autonomous vehicles. While most existing research focuses on improving the accuracy of HD map construction, multi-modal fusion approaches—integrating data from complementary sensors like cameras and LiDAR—have shown promising results by leveraging the strengths of both.

However, in real-world autonomous driving scenarios, perception systems must operate under diverse and often challenging conditions. These include sensor corruptions caused by adverse weather (\textit{e.g.}, snow, fog), sensor failures (\textit{e.g.},  camera crashes, LiDAR misalignment), and external disturbances, all of which can significantly degrade model performance.
Despite these challenges, the robustness of HD map construction models—defined as their ability to sustain performance under such corruptions—has largely been overlooked in previous studies. This oversight creates a significant gap in ensuring the reliability and safety of autonomous driving systems.

\begin{figure*}[!h]
    \centering
    \includegraphics[width=0.92\textwidth]{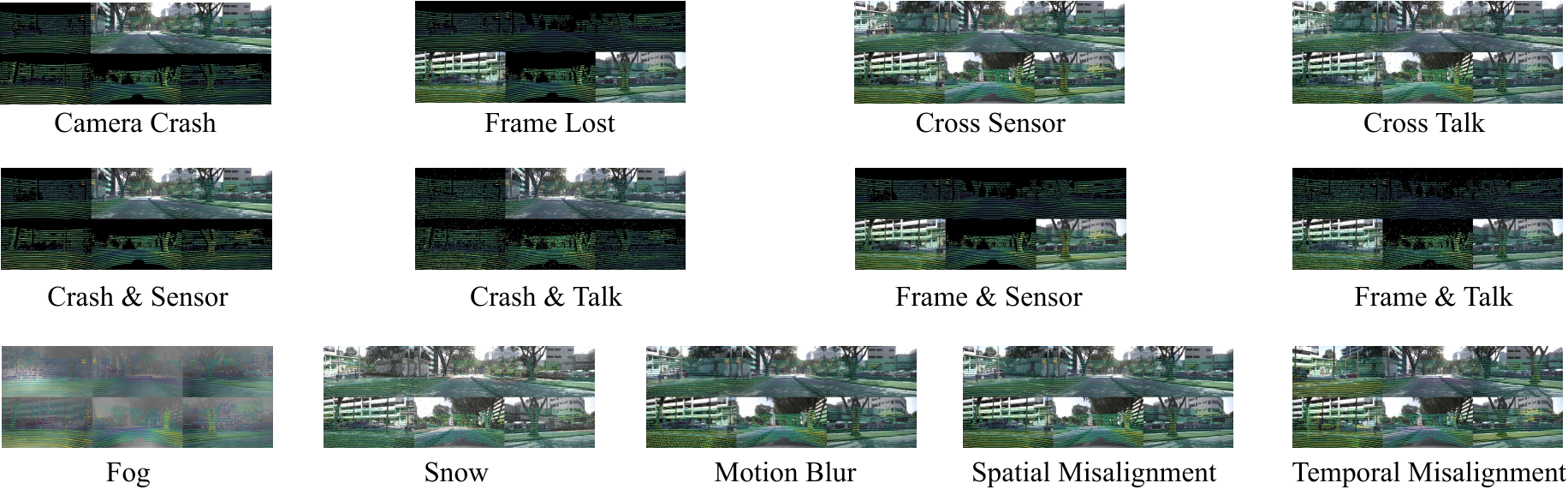}
    \caption{
\textbf{Overview of the Multi-Sensor Corruption dataset.}
 Multi-Sensor Corruption includes 13 types of synthetic camera-LiDAR corruption combinations that perturb both camera and LiDAR inputs, either separately or concurrently.   }
    \label{fig1}
\vspace{-1em}
\end{figure*}

To address this gap, we investigate the robustness of multi-modal fusion methods for HD map construction while maintaining high accuracy. Specifically, we aim to answer two key questions: How do HD map construction models perform under various sensor corruptions, and what strategies can enhance their robustness without compromising accuracy? To achieve this, we propose three key components: data augmentation, a multi-modal fusion module, and training strategies. These components are designed to improve the resilience of HD map construction models against 13 types of multi-sensor corruptions, including both single-source and multi-source disruptions, as illustrated in Fig.~\ref{fig1}.

We evaluate our approach on a \textit{Multi-Sensor Corruption} dataset and benchmark its performance against baseline methods. Experimental results demonstrate that the proposed components significantly enhance the robustness of HD map construction models while achieving state-of-the-art performance on the clean validation set of the NuScenes dataset. These findings provide valuable insights for improving the robustness and reliability of HD map construction models, advancing their applicability in real-world autonomous driving systems.
To summarize, the contributions of this paper are three-fold:

\begin{itemize}
\item \textbf{Comprehensive Robustness Benchmarking}: We conduct a systematic evaluation of multi-modal HD map construction methods using a dataset with 13 types of \textit{Multi-Sensor Corruption}. This provides a thorough analysis of model performance under challenging conditions.

\item 
\textbf{Enhancing Framework}: We propose three key components—data augmentation, a novel multi-modal fusion module, and a modality dropout training strategy—that significantly enhance the robustness of multi-modal fusion methods without sacrificing accuracy.

\item \textbf{State-of-the-Art Performance}: Our approach not only strengthens model resilience against sensor corruptions but also achieves state-of-the-art results on the NuScenes dataset's clean validation set, demonstrating its effectiveness in real-world autonomous driving scenarios.
\end{itemize}

\section{Related Work}

\subsection{HD Map Construction}
The HD map construction task~\cite{hao2024mapdistill,wang2024stream,chen2025stvit+,jia2024diffmap} focuses on generating high-resolution, precise maps that provide instance-level vectorized representations of geometric and semantic elements, such as lane boundaries and road structures. These maps are essential for accurate localization and path planning in autonomous driving systems.
Recent advancements in camera-LiDAR fusion methods~\cite{MapTR,zhao2025fastrsr,hao2025mapfusion,bevfusion22mips,wang2022deepfusionmot} have demonstrated the benefits of combining the semantic richness of camera data with the geometric precision of LiDAR. 
A particularly promising approach is BEV-level fusion, which encodes raw inputs from both sensors into a unified Bird's Eye View (BEV) space. This method effectively integrates complementary features from multiple modalities, achieving superior performance compared to uni-modal approaches.
However, existing methods assume ideal conditions with complete and uncorrupted sensor data, leading to poor robustness in real-world scenarios where data may be missing or compromised. Such reliance on perfect sensor inputs often results in significant performance degradation or complete system failure under adverse conditions. In this paper, we investigate the critical factors necessary for achieving robust multi-sensor HD map construction.

\subsection{Autonomous Driving Perception Robustness}
Recent research has increasingly focused on the robustness of autonomous driving perception tasks~\cite{haosafemap,hao2024team,kong2024robodrive,haousing}. Studies such as RoBoBEV~\cite{xie2023robobev} evaluate the robustness of Bird's Eye View (BEV) perception, while others develop more resilient models or propose strategies to enhance system robustness~\cite{song2024robustness,li2024the}. Robo3D~\cite{kong2023robo3d} benchmarks LiDAR-based semantic segmentation and 3D object detection under conditions of sensor corruption and failure. Zhu et al.~\cite{Zhu_2023_CVPR} assess the natural and adversarial robustness of BEV-based models, introducing a 3D-consistent patch attack to improve spatiotemporal realism in autonomous driving. Additionally, MapBench~\cite{hao2024your} provides benchmarks for evaluating the robustness of HD map construction methods.
In this paper, we investigate the robustness of camera-LiDAR fusion models for HD map construction by designing 13 types of corruption combinations that perturb camera and LiDAR inputs, either individually or simultaneously. Our proposed RoboMap model demonstrates superior robustness across diverse sensor failure scenarios. To the best of our knowledge, RoboMap is the first study to systematically explore the robustness of HD map construction under multi-sensor corruptions.

\begin{figure*}[!h]
    \centering
    \includegraphics[width=1.0\textwidth]{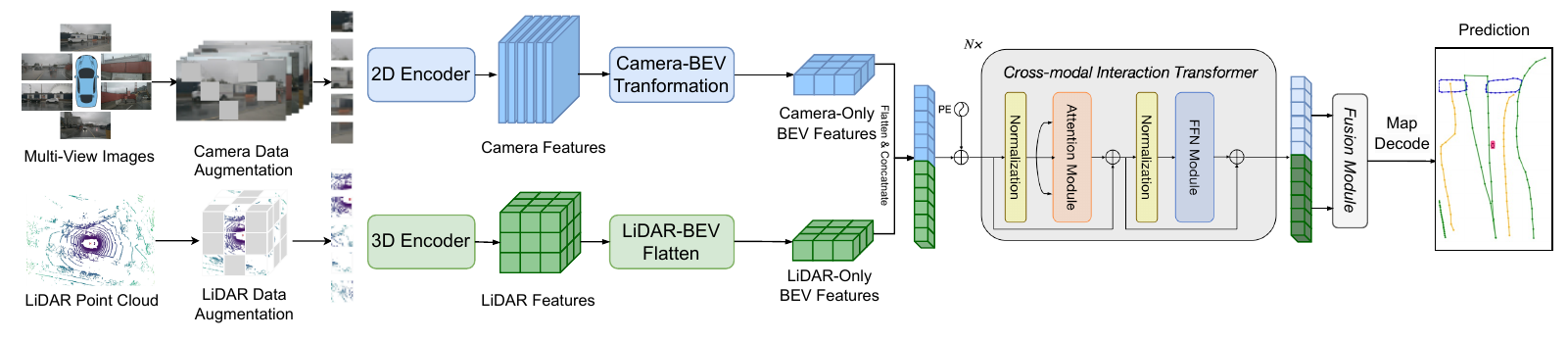}
    \caption{
\textbf{Overview of the RoboMap Framework.} 
The RoboMap framework begins by applying data augmentation to both camera images and LiDAR point clouds. Next, features are efficiently extracted from the multi-modal sensor inputs and transformed into a unified Bird's-Eye View (BEV) space using view transformation techniques. We then introduce a novel multi-modal BEV fusion module to effectively integrate features from both modalities. Finally, the fused BEV features are passed through a shared decoder and prediction heads to generate high-definition (HD) maps.
    }
    \label{fig2}
\end{figure*}

\section{Multi-Sensor Corruption Dataset}

\textbf{Dataset Construction.} In this paper, we investigate the robustness of camera-LiDAR fusion-based HD map construction tasks under various multi-sensor corruptions. Following the protocol established in~\cite{hao2024your,hao2025msc}, we consider three corruption severity levels: \textit{Easy}, \textit{Moderate}, and \textit{Hard}, for each type of corruption. The Multi-Sensor Corruption dataset is constructed by corrupting the \textit{validation} set of the nuScenes dataset~\cite{20cvprnuscense}, which is widely adopted in recent HD map construction research.
The Multi-Sensor Corruption dataset includes $13$ types of synthetic camera-LiDAR corruption combinations, perturbing camera and LiDAR inputs either separately or concurrently, as illustrated in Fig.~\ref{fig1}. These corruptions are categorized into three groups: camera-only, LiDAR-only, and multi-modal corruptions, addressing a wide range of real-world scenarios. Specifically:

\begin{itemize}
    \item \textbf{Camera-Only Corruptions}: We design $2$ types of corruptions using clean LiDAR data to simulate scenarios where the camera system is compromised while the LiDAR remains functional. These include:
    \begin{itemize}
        \item \textit{Camera Crash}: Simulates a complete failure of the camera system, where no visual data is available. This tests the model's ability to rely on LiDAR inputs.
        \item \textit{Frame Lost}: Mimics intermittent camera failures, where certain frames are dropped or missing. This evaluates the model's robustness to visual data.
    \end{itemize}

    \item \textbf{LiDAR-Only Corruptions}: We create $2$ types of corruptions using clean camera data to simulate scenarios where the LiDAR system is compromised while the camera remains operational. These include:
    \begin{itemize}
        \item \textit{Crosstalk}: Simulates interference between LiDAR sensors, where signals from one sensor affect another, leading to noisy or inaccurate point cloud data.
        \item \textit{Cross-Sensor}: Mimics misalignment or calibration errors between LiDAR sensors, resulting in inconsistent or distorted point cloud representations.
    \end{itemize}

    \item \textbf{Multi-Modal Corruptions}: We propose $9$ types of corruptions that perturb both camera and LiDAR inputs to simulate real-world scenarios where both modalities are affected. These include:
    \begin{itemize}
        \item $4$ combinations of the aforementioned failure types (\textit{e.g.}, simultaneous \textit{Camera Crash} and \textit{Crosstalk}), testing the model's resilience to sensor failures.
        \item $5$ additional corruptions:
        \begin{itemize}
            \item \textit{Fog}: Simulates reduced visibility in both camera images and LiDAR point clouds due to dense fog.
            \item \textit{Snow}: Mimics the impact of snowfall, which can obscure camera images and scatter LiDAR signals.
            \item \textit{Motion Blur}: Represents blurring in camera images and distortion in LiDAR data caused by rapid vehicle motion.
            \item \textit{Spatial Misalignment}: Simulates misalignment between camera and LiDAR data due to calibration errors or physical shifts.
          \item \textit{Temporal Misalignment}: Mimics timing discrepancies between camera and LiDAR data, where inputs from the modalities are not synchronized.
        \end{itemize}
    \end{itemize}
\end{itemize}

Using this dataset, we conduct a systematic evaluation of the robustness of multi-modal HD map construction methods, providing a comprehensive analysis of model performance under adverse conditions.

\textbf{Robustness Evaluation Metrics}  
To assess the robustness of HD map construction methods under multi-modal corrupted scenarios, we introduce two evaluation metrics.  

\textit{ Resilience Score (RS)} 
We define $\mathrm{RS}$ as the relative robustness indicator for measuring how much accuracy a model can retain when evaluated on the corruption sets, which are calculated as follows:
\begin{equation}
    \mathrm{RS}_i=\frac{\sum_{l=1}^3\mathrm{Acc}_{i,l}}{3\times\mathrm{Acc}^\mathrm{clean}} ,\quad\mathrm{mRS}=\frac1N\sum_{i=1}^N\mathrm{RS}_i ,
\end{equation}
where $\mathrm{Acc}_{i,l}$ denotes the task-specific accuracy scores, with NDS (NuScenes Detection Score) for 3D object detection and mAP (mean Average Precision) for HD map construction, on corruption type $i$ at severity level $l$. $N$ is the total number of corruption types, and $\mathrm{Acc}^\mathrm{clean}$ denotes the accuracy score on the ``clean'' evaluation set. 
$\mathrm{mRS}$ (mean Resilience Score) represents the average score, providing an overall measure of the model's robustness across all types of corruption.

\textit{Relative Resilience Score (RRS)}
We define $\mathrm{RRS}$ as the critical metric for comparing the relative robustness of candidate models with the baseline model and $\mathrm{mRRS}$ as an overall metric to indicate the relative resilience score. The $\mathrm{RRS}$ and $\mathrm{mRRS}$ scores are calculated as follows:
\begin{equation}
    \mathrm{RRS}_i=\frac{\sum_{l=1}^3\mathrm{Acc}_{i,l}}{\sum_{l=1}^3\mathrm{Acc}_{i,l}^\mathrm{base}} -1,\quad\mathrm{mRRS}=\frac1N\sum_{i=1}^N\mathrm{RRS}_i ,
\end{equation}
where $\mathrm{Acc}_{i,l}^\mathrm{base}$ denotes the accuracy  of the baseline model.

\section{Method}
\textbf{Preliminaries}
For clarity, we first introduce the notation and definitions used throughout this paper. Our goal is to design a robust multi-modal HD map construction framework that integrates data augmentation, a novel multi-modal fusion module, and effective training strategies to significantly enhance the robustness of multi-modal fusion methods, as illustrated in Fig.~\ref{fig2}. Formally, let $\chi = \{\textit{Camera}, \textit{LiDAR}\}$ represent the set of inputs, where $\textit{Camera} \in \mathbb{R}^{B \times N^{cam} \times H^{cam} \times W^{cam} \times 3}$ denotes multi-view RGB camera images in perspective view (with $B$, $N^{cam}$, $H^{cam}$, and $W^{cam}$ representing batch size, number of cameras, image height, and image width, respectively), and $\textit{LiDAR} \in \mathbb{R}^{B \times P \times 5}$ represents the LiDAR point cloud (with $P$ points, each containing 3D coordinates, reflectivity, and beam index). The detailed architectural designs are described in the following sections.

\begin{table*}[t]
\begin{center}
\caption{
\textbf{Comparisons with state-of-the-art methods on
nuScenes val set.}
L'' and C'' represent LiDAR and camera, respectively.
Effi-B0'', R50'', PP'', and Sec'' are short for EfficientNet-B0, ResNet50, PointPillars and SECOND, respectively. Note that RoboMap (MapModel) means our method is integrated into an existing MapModel. Best viewed in color.
}
\scalebox{1.0}{
\begin{tabular}{l|c|cccc|cccc}
\hline
\rowcolor{black!10} \textbf{Method} & \textbf{Venue} & \textbf{Modality} & \textbf{BEV Encoder} & \textbf{Backbone} & \textbf{Epoch} & $\textbf{AP}_{ped.}$ & $\textbf{AP}_{div.}$ & $\textbf{AP}_{bou.}$ & $\textbf{mAP} \uparrow$ \\
\midrule
\makecell[l]{HDMapNet\cite{li2022hdmapnet}} & ICRA'22 & \makecell[c]{C} & \makecell[c]{NVT} & \makecell[c]{Effi-B0} & \makecell[c]{30} & \makecell[c]{14.4} & \makecell[c]{21.7} & \makecell[c]{33.0} & \makecell[c]{23.0} \\

\makecell[l]{VectorMapNet~\cite{liu2023vectormapnet}} & ICML'23 & \makecell[c]{C} & \makecell[c]{IPM} & \makecell[c]{R50} & \makecell[c]{110} & \makecell[c]{36.1} & \makecell[c]{47.3} & \makecell[c]{39.3} & \makecell[c]{40.9} \\

\makecell[l]{PivotNet~\cite{ding2023pivotnet}} & ICCV'23 & \makecell[c]{C} & \makecell[c]{PersFormer} & \makecell[c]{R50} & \makecell[c]{30} & \makecell[c]{53.8} & \makecell[c]{58.8} & \makecell[c]{59.6} & \makecell[c]{57.4} \\

\makecell[l]{BeMapNet~\cite{qiao2023end}} & CVPR'23 & \makecell[c]{C} & \makecell[c]{IPM-PE} & \makecell[c]{R50} & \makecell[c]{30} & \makecell[c]{57.7} & \makecell[c]{62.3} & \makecell[c]{59.4} & \makecell[c]{59.8} \\

\makecell[l]{MapVR~\cite{zhang2023online}} & NeurIPS'24 & \makecell[c]{C} & \makecell[c]{GKT} & \makecell[c]{R50} & \makecell[c]{24} & \makecell[c]{47.7} & \makecell[c]{54.4} & \makecell[c]{51.4} & \makecell[c]{51.2} \\

\makecell[l]{MapTRv2~\cite{maptrv2}} & IJCV'24 & \makecell[c]{C} & \makecell[c]{BEVPoolv2} & \makecell[c]{R50} & \makecell[c]{24} & \makecell[c]{59.8} & \makecell[c]{62.4} & \makecell[c]{62.4} & \makecell[c]{61.5} \\

StreamMapNet~\cite{yuan2024streammapnet} & WACV'24 & {C} & BEVFormer & {R50} & {$30$} & {$61.7$} & {$66.3$} & {$62.1$} & {$63.4$} \\

MapTR~\cite{MapTR} & ICLR'23 & C & GKT & R50 & 24 & 46.3 & 51.5 & 53.1 & 50.3 \\

HIMap~\cite{HIMap} & CVPR'24 & C & BEVFormer & R50 & 24 & 62.2 & 66.5 & 67.9 & 65.5 \\

\hline
\hline
\makecell[l]{VectorMapNet~\cite{liu2023vectormapnet}} & ICML'23 & \makecell[c]{L} & \makecell[c]{-} & \makecell[c]{PP} & \makecell[c]{110} & \makecell[c]{25.7} & \makecell[c]{37.6} & \makecell[c]{38.6} & \makecell[c]{34.0} \\

\makecell[l]{MapTRv2~\cite{maptrv2}} & IJCV'24 & \makecell[c]{L} & \makecell[c]{-} & \makecell[c]{Sec} & \makecell[c]{24} & \makecell[c]{56.6} & \makecell[c]{58.1} & \makecell[c]{69.8} & \makecell[c]{61.5} \\

MapTR~\cite{MapTR} & ICLR'23 & L & - & Sec & 24 & 48.5 & 53.7 & 64.7 & 55.6 \\

HIMap~\cite{HIMap} & CVPR'24 & L & - & Sec & 24 & 54.8 & 64.7 & 73.5 & 64.3 \\

\hline
\midrule
HDMapNet~\cite{li2022hdmapnet} & ICRA'22 & C \& L & NVT & Effi-B0 \& PP & 30 & 16.3 & 29.6 & 46.7 & 31.0 \\

VectorMapNet~\cite{liu2023vectormapnet} & ICML'23 & C \& L & IPM & R50 \& PP & 110+ft & 48.2 & 60.1 & 53.0 & 53.7 \\

MBFusion~\cite{hao2024mbfusion} & ICRA'24 & C \& L & GKT & R50 \& Sec & 24 & 61.6 & 64.4 & 72.5 & 66.1 \\

GeMap~\cite{GeMap} & ECCV'24 & C \& L & GKT & R50 \& Sec & 24 & 66.3 & 62.2 & 71.1 & 66.5 \\

MapTRv2~\cite{maptrv2} & IJCV'24 & C \& L & BEVPoolv2 & R50 \& Sec & 24 & 65.6 & 66.5 & 74.8 & 69.0 \\

Mgmap~\cite{Mgmap} & CVPR'24 & C \& L & GKT & R50 \& Sec & 24 & 67.7 & 71.1 & 76.2 & 71.7 \\

\rowcolor{black!10} MapTR~\cite{MapTR} & ICLR'23 & C \& L & GKT & R50 \& Sec & 24 & 55.9 & 62.3 & 69.3 & 62.5 \\

\rowcolor{black!10} HIMap~\cite{HIMap} & CVPR'24 & C \& L & BEVFormer & R50 \& Sec & 24 & 71.0 & 72.4 & 79.4 & 74.3 \\

\rowcolor{green!10} \textbf{RoboMap (MapTR)} & - & \textbf{C \& L} & \textbf{GKT} & \textbf{R50 \& Sec} & \textbf{24} & \textbf{67.8} & \textbf{70.4} & \textbf{76.4} & \textbf{71.5} \\

\rowcolor{green!10} \textbf{RoboMap (HIMap)} & - & \textbf{C \& L} & \textbf{BEVFormer} & \textbf{R50 \& Sec} & \textbf{24} & \textbf{74.6} & \textbf{74.5} & \textbf{82.0} & \textbf{77.0} \\

\hline
\bottomrule
\end{tabular}}
\label{tab1}
\end{center}
\end{table*}

\textbf{Data Augmentation}  
To enhance robustness against sensor corruptions, we employ data augmentation strategies for both camera and LiDAR inputs. For camera data, we utilize GridMask~\cite{shorten2019survey}, which randomly drops image information by applying a grid mask of the same size as the image, with binary values (0 or 1). For LiDAR data, we apply a dropout strategy~\cite{choi2021part} that randomly removes points from the point cloud to simulate sensor noise and improve model resilience.

After augmentation, we process the data as follows: For Camera Data, we utilize ResNet50~\cite{he2016deep} as the backbone to extract multi-view features and apply GKT~\cite{2022GKT} as the 2D-to-BEV transformation module, converting these features into Bird's-Eye View (BEV) space. This results in BEV features \( F_{Camera}^{BEV} \in \mathbb{R}^{B\times H\times W\times C} \), where \( H \), \( W \), and \( C \) denote height, width, and number of channels, respectively. For LiDAR Data, we follow the SECOND method~\cite{2018seoncd} for voxelization and sparse LiDAR encoding. The resulting LiDAR features are projected into BEV space using a flattening operation as described in~\cite{liu2023bevfusion}, yielding a unified LiDAR BEV representation \( F_{LiDAR}^{BEV} \in \mathbb{R}^{B\times H\times W\times C} \).

\textbf{Cross-modal Interaction Transform}
Existing methods convert sensory features into a shared BEV representation and fuse them to create multi-modal BEV features. However, LiDAR and camera features remain semantically misaligned due to modality gaps. To address this, we propose a Cross-Modal Interaction Transformer (CIT) module utilizing self-attention to enrich one modality with insights from another.

First, we start with the BEV features from both the camera ($F_{Camera}^{BEV} \in \mathbb{R}^{B\times H\times W\times C}$) and LIDAR ($F_{LiDAR}^{BEV} \in \mathbb{R}^{B\times H\times W\times C}$) sensors. The BEV tokens $\mathbf{T}_{\mathrm{Camera}}^{\mathrm{BEV}} \in \mathbb{R}^{HW\times C}$
and $\mathbf{T}_{\mathrm{LiDAR}}^{\mathrm{BEV}} \in \mathbb{R}^{HW\times C}$ are obtained by flattening each BEV feature and permuting the order of the matrices.
Next, we concatenate the tokens of each modality and add a learnable positional embedding, which is a trainable parameter of dimension $2HW\times C$, 
to create the input BEV tokens $\mathbf{T}^{\mathrm{in}} \in \mathbb{R}^{2HW\times C}$ for the Transformer. This positional embedding allows the model to distinguish spatial information between different tokens during training.
Third, the input tokens $\mathbf{T}^{\mathrm{in}}$ undergo linear projections to compute a set of queries, keys, and values ($\mathbf{Q}$, $\mathbf{K}$ and $\mathbf{V}$).
Fourth, the self-attention layer computes the attention weights using scaled the dot product between 
$\mathbf{Q}$ and $\mathbf{K}$, and then multiplies these weights by the values to produce the refined output,
\begin{equation} \label{eq2}
\mathbf{Z} = \operatorname{Attention}(\mathbf{Q}, \mathbf{K}, \mathbf{V}) = \operatorname{softmax}\left(\frac{\mathbf{Q}\mathbf{K}^T}{\sqrt{D_\mathrm{k}}}\right)\mathbf{V},
\end{equation}
where 
$\frac{1}{\sqrt{D_\mathrm{k}}}$
is a scaling factor.
To capture complex relationships across various representation subspaces and positions, we adopt the multi-head attention mechanism,
\begin{equation} \label{eq3}
\begin{aligned}
  \hat{\mathbf{Z}} &= \operatorname{MultiHead}(\mathbf{Q}, \mathbf{K}, \mathbf{V}) = \operatorname{Concat}(\mathbf{Z}_1, \cdots, \mathbf{Z}_h)\mathbf{W}^\mathrm{O}.
\end{aligned}
\end{equation}
The subscript $h$ denotes the number of head, and
$\mathbf{W}^\mathrm{O}$
denotes the projected matrix of 
$\operatorname{Concat}(\mathbf{Z}_1, \cdots, \mathbf{Z}_h)$.
Finally, the transformer uses a non-linear transformation to
calculate the output features, 
$\mathbf{T}^{\mathrm{out}}$ 
which are of the same shape as the input features 
$\mathbf{T}^{\mathrm{in}}$,
\begin{equation} \label{eq4}
\mathbf{T}^{\mathrm{out}} = \operatorname{MLP}(\hat{\mathbf{Z}}) + \mathbf{T}^{\mathrm{in}}.
\end{equation}
The output 
$\mathbf{T}^{\mathrm{out}}$ 
are converted into 
$\hat{\mathbf{F}}_{\mathrm{Camera}}^{\mathrm{BEV}}$ and $\hat{\mathbf{F}}_{\mathrm{LiDAR}}^{\mathrm{BEV}}$
for further feature fusion. 
We utilize the Dynamic Fusion module to aggregate the multi-modal BEV feature inputs, $\hat{\mathbf{F}}_{\mathrm{Camera}}^{\mathrm{BEV}}$ and $\hat{\mathbf{F}}_{\mathrm{LiDAR}}^{\mathrm{BEV}}$, resulting in the aggregated features $\mathbf{F}_{\mathrm{fused}}$.
The output  fused  feature $\textbf{F}_{fused}$ will be used for HD Map construction task, with the decoder and prediction heads.

\textbf{Modality Dropout Training Strategy}
To simulate real-world sensor failures during training, we employ a Modality Dropout strategy, where the BEV features of either the camera or LiDAR ($\hat{\mathbf{F}}_{\mathrm{Camera}}^{\mathrm{BEV}}$ or $\hat{\mathbf{F}}_{\mathrm{LiDAR}}^{\mathrm{BEV}}$) are randomly dropped with a probability \( p_{\mathrm{md}} \). When a modality is dropped, \( p_{\mathrm{L}} \) denotes the probability of retaining the LiDAR input, while \( p_{\mathrm{C}} = 1 - p_{\mathrm{L}} \) represents the probability of retaining the camera input. Thus, the overall probability distribution is as follows: the probability of retaining both sensors is \( 1 - p_{\mathrm{md}} \), the probability of retaining only LiDAR is \( p_{\mathrm{md}} \cdot p_{\mathrm{L}} \), and the probability of retaining only the camera is \( p_{\mathrm{md}} \cdot (1 - p_{\mathrm{L}}) \). This strategy enhances the model's robustness to partial sensor failures by randomly dropping modalities, enabling it to better adapt to real-world scenarios where sensor malfunctions may occur.

\begin{table*}[t]
\centering
\caption{
    \textbf{The scores $RS_{c}$ and $mRS$  for the original MapTR~\cite{MapTR} model and its variants. $RS_{c}$ using mAP as metric.} 
}
\vspace{-0.1cm}
\scalebox{0.62}{
    \begin{tabular}{l|ccccccccccccc|c}
    \toprule
    \textbf{Model}& \begin{tabular}[c]{@{}c@{}}\textbf{Motion} \\ \textbf{Blur}\end{tabular} & \begin{tabular}[c]{@{}c@{}}\textbf{Temporal} \\ \textbf{Mis.}\end{tabular} & \begin{tabular}[c]{@{}c@{}}\textbf{Spatial} \\ \textbf{Mis.}\end{tabular} & \textbf{Fog} & \textbf{Snow} & \textbf{Camera Crash}& \textbf{Frame Lost} &  \textbf{Cross Sensor} &  \textbf{Cross Talk}  & \begin{tabular}[c]{@{}c@{}}\textbf{Camera Crash,} \\ \textbf{Cross Sensor}\end{tabular} & \begin{tabular}[c]{@{}c@{}}\textbf{Camera Crash,} \\ \textbf{Cross Talk}\end{tabular}  & \begin{tabular}[c]{@{}c@{}}\textbf{Frame Lost,} \\ \textbf{Cross Sensor}\end{tabular} & \begin{tabular}[c]{@{}c@{}}\textbf{Frame Lost,} \\ \textbf{Cross Talk}\end{tabular}  & \begin{tabular}[c]{@{}c@{}}\textbf{mRS}$\uparrow$\end{tabular}  
    \\\midrule\midrule
   MapTR (Baseline)& 70.00  & 76.94  & 69.05 & 67.94  & 19.55  & 78.69  & 74.75  & 98.47  & 84.99  & 77.40  & 58.14  & 73.50  & 54.27  & 69.51  
   \\
MapTR (Baseline) + Fusion Module  & 80.03  & 75.28  & 68.48  & 68.26  & 23.67  & 70.77  & 64.15  & 96.34  & 87.93  & 69.52  & 56.57  & 62.94  & 51.03  & 67.31            \\
MapTR (Baseline) + Data Augmentation  & 71.70  & 75.04  & 68.16  & 66.76  & 24.66  & 79.29  & 76.88  & 96.63  & 90.31  & 77.81  & 66.92  & 75.42  & 63.86  & 71.80         \\
MapTR (Baseline) + Dropout Training & 72.11  & 73.52  & 57.10  & 63.19  & 20.59  & 82.23  & 80.55  & 94.09  & 81.13  & 80.34  & 64.06  & 78.67  & 62.00  & 69.97            \\
\textbf{RoboMap (MapTR)} & \textbf{89.88}  & \textbf{73.48}  & \textbf{63.02}  & \textbf{69.17}  &  \textbf{23.78}  & \textbf{93.49}  & \textbf{92.86}  & \textbf{95.90}  & \textbf{86.83}  & \textbf{91.85}  & \textbf{78.95}  & \textbf{91.07}  & \textbf{77.56}  & \textbf{79.06}                               
    \\\bottomrule
    \end{tabular}}
    \label{tab2}
\end{table*}
\begin{table*}[t]
\centering
\caption{
    \textbf{The scores $RS_{c}$ and $mRS$  for the original HIMap~\cite{HIMap} model and its variants. $RS_{c}$ using mAP as metric.} 
}
\vspace{-0.1cm}
\scalebox{0.62}{
    \begin{tabular}{l|ccccccccccccc|c}
    \toprule
    \textbf{Model} & \begin{tabular}[c]{@{}c@{}}\textbf{Motion} \\ \textbf{Blur}\end{tabular} & \begin{tabular}[c]{@{}c@{}}\textbf{Temporal} \\ \textbf{Mis.}\end{tabular} & \begin{tabular}[c]{@{}c@{}}\textbf{Spatial} \\ \textbf{Mis.}\end{tabular} & \textbf{Fog} & \textbf{Snow} & \textbf{Camera Crash}& \textbf{Frame Lost} &  \textbf{Cross Sensor} &  \textbf{Cross Talk}  & \begin{tabular}[c]{@{}c@{}}\textbf{Camera Crash,} \\ \textbf{Cross Sensor}\end{tabular} & \begin{tabular}[c]{@{}c@{}}\textbf{Camera Crash,} \\ \textbf{Cross Talk}\end{tabular}  & \begin{tabular}[c]{@{}c@{}}\textbf{Frame Lost,} \\ \textbf{Cross Sensor}\end{tabular} & \begin{tabular}[c]{@{}c@{}}\textbf{Frame Lost,} \\ \textbf{Cross Talk}\end{tabular}  & \begin{tabular}[c]{@{}c@{}}\textbf{mRS}$\uparrow$\end{tabular}  
    \\\midrule\midrule
   HIMap (Baseline) & 83.77  & 74.93  & 77.31  & 75.56  & 23.79  & 63.41  & 58.59  & 97.84  & 94.28  & 61.91  & 54.15  & 57.19  & 57.19  & 67.69            \\
HIMap (Baseline) + Fusion Module  & 83.69  & 72.92  & 79.06  & 75.17  & 25.59  & 68.58  & 63.92  & 97.69  & 94.86  & 67.15  & 60.57  & 62.56  & 56.16  & 69.84           \\
HIMap (Baseline) + Data Augmentation& 84.35  & 73.30  & 79.21  & 75.44  & 25.50  & 66.27  & 61.11  & 97.52  & 94.93  & 64.81  & 58.63  & 59.64  & 53.63  & 68.80  \\
HIMap (Baseline) + Dropout Training& 84.21  & 72.09  & 71.73  & 70.93  & 27.13  & 87.41  & 85.15  & 96.49  & 91.50  & 85.69  & 76.56  & 83.39  & 72.92  & 77.32             \\
\textbf{RoboMap (HIMap)} & \textbf{90.34}  & \textbf{72.54}  & \textbf{72.32}  & \textbf{76.69}  & \textbf{30.44}  & \textbf{93.17} & \textbf{92.58}  & \textbf{97.21}  & \textbf{93.24}  & \textbf{91.68} & \textbf{83.26}  & \textbf{90.95}  & \textbf{81.43}  & \textbf{81.99}  
    \\\bottomrule
    \end{tabular}}
    \label{tab3}
\end{table*}

\section{Experiments and Analysis}
\subsection{Experimental Settings}
\textbf{Dataset}
The nuScenes dataset~\cite{20cvprnuscense} consists of 1,000 sequences collected by autonomous vehicles. Each sample is annotated at 2Hz and includes six camera images capturing the $360^\circ$ horizontal field of view of the ego-vehicle. Following the methodologies in~\cite{MapTR,HIMap}, we focus on three key map elements: pedestrian crossings, lane dividers, and road boundaries, to ensure a fair evaluation.

\textbf{Evaluation Metrics}
For clean data, we adopt metrics consistent with prior HD map studies~\cite{MapTR,xiaoshuai2025electronic,HIMap}. Average Precision (AP) measures the quality of map construction, while \textbf{Chamfer Distance ($D_{\text{Chamfer}}$)} quantifies the alignment between predictions and ground truth. To assess model robustness, we introduce the Resilience Score (RS) and Relative Resilience Score (RRS), which evaluate the model's performance under data corruption or sensor noise, ensuring reliability in real-world scenarios.

\textbf{Implementation Details}
Our RoboMap framework is trained on four NVIDIA RTX A6000 GPUs. We retrain two state-of-the-art baseline models, MapTR~\cite{MapTR} and HIMap~\cite{HIMap}, using their official configurations from open-source repositories. All experiments employ the AdamW optimizer with a learning rate of $4.2 \times 10^{-4}$. Notably, RoboMap's core components—data augmentation, multi-modal fusion module, and training strategies—are designed as simple yet effective plug-and-play techniques, making them compatible with existing camera-LiDAR fusion pipelines for HD map construction.

\begin{figure}[t]
    \centering
\setlength{\abovecaptionskip}{-0.01cm}  
\includegraphics[width=1.0\linewidth]{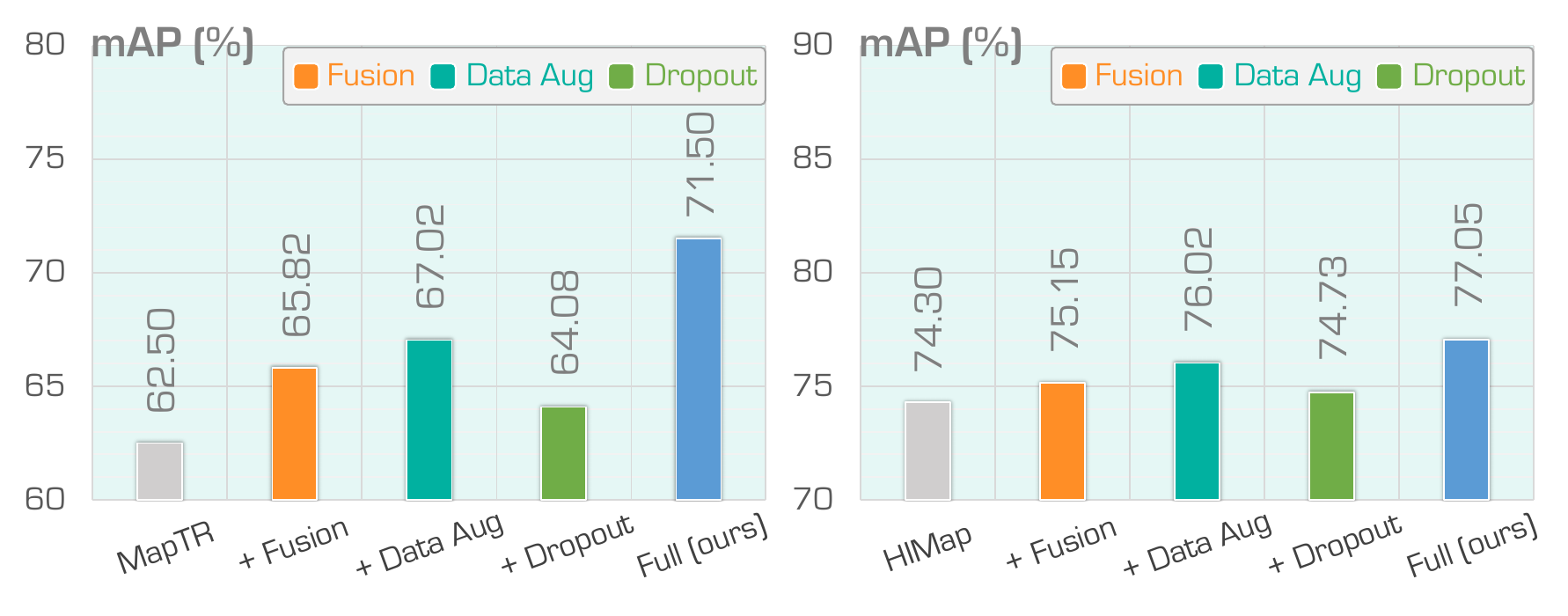}
    \caption{
 Analyze the impact of different modules on the HD map construction task using clean data.}
    \label{abla}
\end{figure}

\begin{figure*}[h]
    \centering
     \setlength{\abovecaptionskip}{-0.01cm}  
     \includegraphics[width=1.0\textwidth]{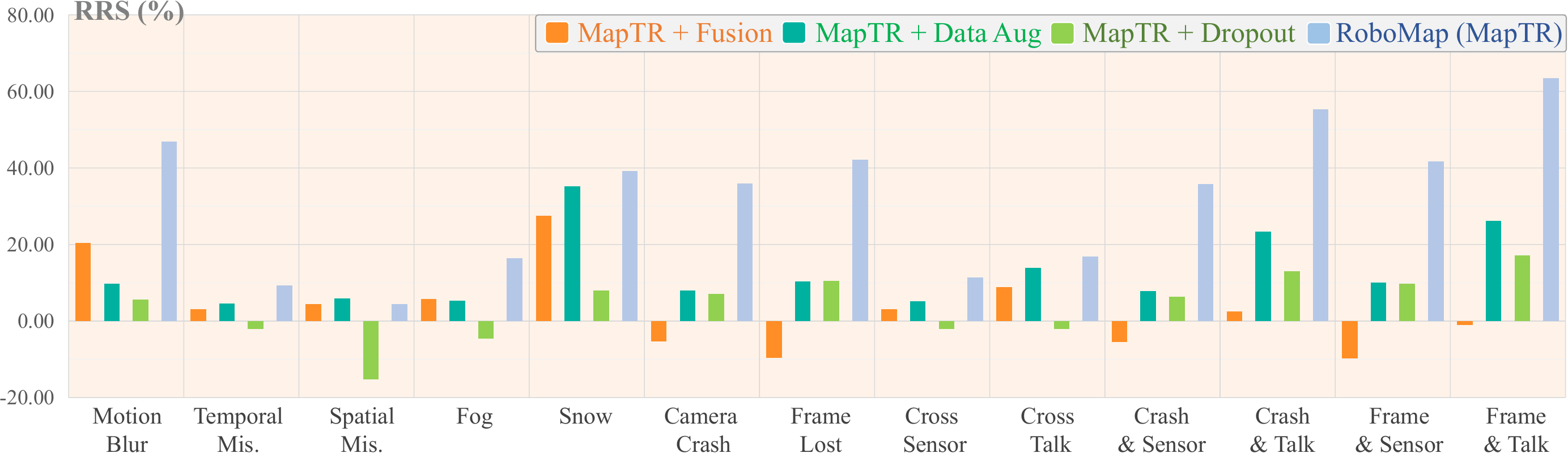}
    \caption{
Relative robustness visualization. Relative Resilience Score (RRS) computed with mAP using original MapTR~\cite{MapTR} as baseline. 
    }
    \label{fig4}
\end{figure*}

\begin{figure*}[h]
    \centering
     \setlength{\abovecaptionskip}{-0.01cm}  
     \includegraphics[width=1.0\textwidth]{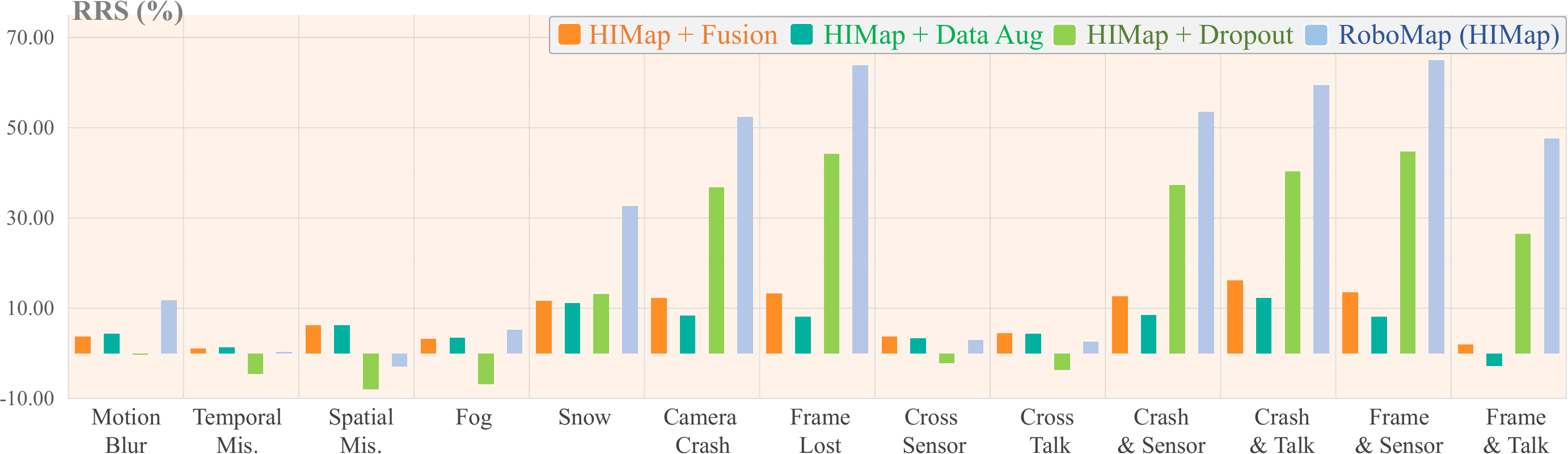}
    \caption{
Relative robustness visualization. Relative Resilience Score (RRS) computed with mAP using original HIMap~\cite{HIMap} as baseline. 
    }
    \label{fig5}
\end{figure*}

\subsection{Comparison with the State-of-the-Arts}
With the same settings and data partition, we compare the proposed RoboMap model with several state-of-the-art methods, including HDMapNet~\cite{li2022hdmapnet}, VectorMapNet~\cite{liu2023vectormapnet}, MBFusion~\cite{hao2024mbfusion}, GeMap~\cite{GeMap}, MgMap~\cite{Mgmap}, MapTR~\cite{MapTR}, MapTRv2~\cite{maptrv2}, and HIMap~\cite{HIMap}. The overall performance of RoboMap and all baselines on the nuScenes dataset is summarized in Tab.~\ref{tab1}. 

The experimental results highlight several key observations: multi-modal approaches consistently outperform single-modal methods, demonstrating the importance of leveraging complementary information from both camera and LiDAR sensors for HD map construction. As shown in Tab.~\ref{tab1}, RoboMap achieves significant improvements over the original models, with RoboMap (MapTR) surpassing the original camera-LiDAR fusion MapTR model by 9 mAP on the nuScenes dataset and RoboMap (HIMap) outperforming the previous state-of-the-art HIMap fusion model by 2.7 mAP, setting a new benchmark for vectorized map reconstruction. The superior performance of RoboMap can be attributed to its three core components—data augmentation, a multi-modal fusion module, and advanced training strategies—which collectively enhance robustness and accuracy. In summary, RoboMap demonstrates substantial superiority over existing multi-modal methods, highlighting its effectiveness in HD map construction tasks.

\subsection{Ablation Studies}
To systematically evaluate the effectiveness of each component in our proposed RoboMap, we conduct ablation studies by incrementally adding individual strategies to the baseline model and present the results in Fig.~\ref{abla}. Specifically, we design the following ablation models: (1) \textbf{RoboMap (w/o Fusion)}, which integrates a cross-modal interaction transformation fusion module into the original baseline model; (2) \textbf{RoboMap (w/ Data Augmentation)}, which incorporates image and LiDAR data augmentation strategies into the original baseline model; (3) \textbf{RoboMap (w/ Dropout)}, which applies the Modality Dropout Training Strategy to the original baseline model; and (4) \textbf{RoboMap (full)}, which combines all three key components—data augmentation, a multi-modal fusion module, and training strategies—into the baseline model.

The ablation results demonstrate that each component significantly enhances the baseline model's performance. Specifically, RoboMap (w/ Fusion), RoboMap (w/ Data Augmentation), and RoboMap (w/ Dropout) outperform the baseline MapTR model on the nuScenes dataset, achieving gains of 3.3, 4.5, and 1.6 mAP, respectively. Similarly, these variants surpass the state-of-the-art HIMap model, with improvements of 0.85, 1.7, and 0.4 mAP, respectively. These extensive experimental results validate the effectiveness of each strategy in improving model performance, highlighting the robustness and versatility of RoboMap.

\subsection{Robustness of multi-sensor corruptions}

To explore strategies that enhance robustness, such as data augmentation, multi-modal fusion, and modality dropout training, we evaluated the popular MapTR~\cite{MapTR} and the state-of-the-art HIMap~\cite{HIMap} models. Tab.~\ref{tab2} and Tab.~\ref{tab3} present their Resilience Scores, while Fig.~\ref{fig4} and Fig.~\ref{fig5} illustrate their Relative Resilience Scores. Our analysis reveals two key insights. First, while camera-LiDAR fusion methods show promising performance by integrating multi-modal data, many approaches assume complete sensor availability, leading to low robustness when sensors are corrupted or missing. Second, although individual strategies do not consistently improve robustness across all multi-sensor corruption scenarios, combining them significantly enhances model resilience. Specifically, our approach improves the mRS metric by 9.55 and 14.3 compared to the original MapTR and HIMap models, respectively, demonstrating the effectiveness of these strategies in boosting robustness.

The experimental results emphasize the need to address sensor vulnerabilities in multi-modal systems. While camera-LiDAR fusion performs well under ideal conditions, its dependence on complete sensor data makes it prone to failure in real-world scenarios with incomplete or corrupted data. By incorporating data augmentation, multi-modal fusion, and modality dropout training, we significantly improve robustness. These strategies enhance the resilience of both MapTR and HIMap models and offer a framework for building more robust multi-modal systems. The findings highlight the potential of targeted enhancements to tackle real-world challenges in sensor-based applications.

\section{Conclusion}
In this paper, we improve the robustness of HD map construction methods, essential for autonomous driving systems. We propose a comprehensive framework integrating data augmentation, a multi-modal fusion module, and innovative modality dropout training strategies. Experimental results demonstrate our method significantly enhances robustness on a dataset with 13 types of sensor corruption. Additionally, our approach achieves state-of-the-art performance on the clean dataset. Overall, our model offers valuable insights for developing more reliable HD map techniques, contributing to safer and more effective autonomous driving technologies.

 \clearpage
\bibliographystyle{IEEEtran}
\bibliography{references}
\end{document}